\documentclass{amia}

\usepackage{algorithm}
\usepackage{algcompatible}
\usepackage{algpseudocode}
\usepackage{amsfonts}
\usepackage{amsmath}
\usepackage{amssymb}
\usepackage{graphicx}
\usepackage{textcomp}
\usepackage{xcolor}

\algnewcommand\algorithmicforeach{\textbf{for each}}
\algdef{S}[FOR]{ForEach}[1]{\algorithmicforeach\ #1\ \algorithmicdo}

\begin{document}

\title{Few-Shot Learning for Clinical Natural Language Processing Using Siamese Neural Networks}

\author{David Oniani$^1$, Sonish Sivarajkumar$^2$, Yanshan Wang$^{1,2,3}$\vspace{4pt}}
\institutes{$^1$Department of Health Information Management, $^2$Intelligence Systems Program, $^3$Department of Biomedical Informatics, University of Pittsburgh, Pittsburgh, PA}


\maketitle


\section*{Abstract}

Clinical Natural Language Processing (NLP) has become an emerging technology in healthcare that
leverages a large amount of free-text data in electronic health records (EHRs) to improve patient
care, support clinical decisions, and facilitate clinical and translational science research.
Recently, deep learning has achieved state-of-the-art performance in many clinical NLP tasks.
However, training deep learning models usually requires large annotated datasets, which are normally
not publicly available and can be time-consuming to build in clinical domains. Working with smaller
annotated datasets is typical in clinical NLP and therefore, ensuring that deep learning models
perform well is crucial for the models to be used in real-world applications. A widely adopted
approach is fine-tuning existing Pre-trained Language Models (PLMs), but these attempts fall short
when the training dataset contains only a few annotated samples. Few-Shot Learning (FSL) has
recently been investigated to tackle this problem. Siamese Neural Network (SNN) has been widely
utilized as an FSL approach in computer vision, but has not been studied well in NLP. Furthermore,
the literature on its applications in clinical domains is scarce. In this paper, we propose two
SNN-based FSL approaches for clinical NLP, including Pre-Trained SNN (PT-SNN) and SNN with
Second-Order Embeddings (SOE-SNN).  We evaluated the proposed approaches on two clinical tasks,
namely clinical text classification and clinical named entity recognition. We tested three few-shot
settings including 4-shot, 8-shot, and 16-shot learning. Both clinical NLP tasks were benchmarked
using three PLMs, including Bidirectional Encoder Representations from Transformers (BERT),
Bidirectional Encoder Representations from Transformers for Biomedical Text Mining (BioBERT), and
Bio \(+\) Clinical BERT (BioClinicalBERT). The experimental results verified the effectiveness of
the proposed SNN-based FSL approaches in both NLP tasks.


\section*{Introduction}

Deep Neural Networks (DNNs), due to their performance \cite{1}, currently dominate both Computer
Vision (CV) and Natural Language Processing (NLP) literature. However, fully utilizing the
capabilities of DNNs requires large training datasets. Reserchers have also tried to reduce the
complexity of the DNN models to obtain comparable performance when the size of training dataset is
small \cite{2}. The Few-Shot Learning (FSL) paradigm is an alternative attempt to tackle the problem
with scarce training instances. The goal of FSL is to efficiently learn from a small number of
``shots'' (i.e., data samples or instances). The number of samples usually ranges from 1 to 100 per
class \cite{6} \cite{7}. There is a growing interest in the Artificial Intelligence (AI) research
community in FSL and several different strategies have been developed for FSL, including Bowtie
Networks \cite{3}, Induction Networks \cite{4}, and Prototypical Networks \cite{5}.

A Siamese Neural Network (SNN), sometimes called a twin neural network, is an Artificial Neural
Network (ANN) that uses two parallel, weight-sharing machine learning models in order to compute
comparable embeddings. The SNN architecture has shown good results as an FSL approach in computer
vision for similarity detection \cite{11} and duplicate identification \cite{12}. Yet, its usage in
NLP has been understudied and, to the best of our knowledge, there have not been any studies
investigating SNNs for clinical NLP.

In SNNs, neural networks need to be trained to compute embeddings. In NLP, deep learning has
achieved the state-of-the-art performance since it could generate comprehensive embeddings to encode
both semantic and syntactic information. The main use of deep learning in NLP is to represent the
language in a vectorized form (i.e., embeddings) so that the vector representation can be used for
different NLP tasks, for example, natural language generation, text classification, and semantic
textual similarity. Thus, embeddings play a key role in applying deep learning to NLP. Having a
robust embedding-generation mechanism is crucial for most NLP tasks. Since the context of words,
sentences, and more generally, text is important to learn meaningful embeddings, context-aware
embedding-generation models, such as BERT \cite{8}, often show promising results. Furthermore,
depending on the domain, the context also varies. For this purpose, engineers and researchers have
built domain-specific, specialized models to be used for downstream tasks. Examples of such models
include BioBERT \cite{9} trained from biomedical literature texts and BioClinicalBERT trained from
clinical texts \cite{10}. Leveraging these contextual embeddings for FSL has rarely been studied in
clinical NLP.

FSL is critical for clinical NLP as annotating a large training dataset is costly. Furthermore, such
annotation often requires involving domain experts. It is not uncommon to have a few clinical text
samples annotated by physicians. One example could be clinical notes with annotations of a rare
disease with the number of samples naturally limited due to the nature of the disease. Despite such
challenges in the clinical domain, the importance of using AI in clinical applications cannot be
understated. AI could not only assist physicians in their decision-making and facilitate clinical
and translational research, but also significantly reduce the need for manual work. Therefore, in
this study, we propose an FSL approach based on SNNs to tackle clinical NLP tasks with only a few
annotated training samples. Two SNN-based FSL approaches have been proposed, including Pre-Trained
SNN (PT-SNN) and SNN with Second-Order Embeddings (SOE-SNN). For every approach, three different
transformer models -- BERT, BioBERT, BioClinicalBERT -- have been utilized. We evaluated the
proposed approaches on two clinical tasks, namely clinical sentence classification and clinical
named entity recognition. Clinical text classification refers to the classification of clinical
sentences based on pre-defined classes. Named Entity Recognition (NER) is a subtask of Information
Extraction (IE) that seeks to identify entities mentioned in clinical texts. We show that SNN-based
approaches outperform the baseline models in few-shot settings for both tasks. Finally, we discuss
the limitations and future work.

\section*{Background and Related Work}

There have been studies evaluating the usability of SNNs for image classification. Mahajan et al.
used SNNs for the classification of high-dimensional radiomic features extracted from MRI images
\cite{13}. Hunt et al. applied SNNs for the classification of electrograms \cite{14}. Zhao et al.
have utilized SNNs for hyperspectral image classification \cite{15}. 

In the context of FSL, SNNs have been used by Torres et al. for one-shot, Convolutional Neural
Networks (CNNs) based classification in order to optimize the discovery of novel compounds based on
a reduced set of candidate drugs \cite{16}. Droghini et al. employed SNNs for few-shot human fall
detection purposes using images \cite{17}. However, none of these studies used SNN-based FSL for
NLP.

There is only a recent study by Müller et al \cite{18} that explored SNNs for FSL in NLP and
demonstrated the high performance of pretrained SNNs that embed texts and labels. To the best of our
knowledge, none of the studies referenced above are using SNNs to perform FSL in the clinical NLP
domain.


\section*{Materials}

We perform both clinical text classification and clinical named entity recognition. For few-shot
clinical text classification, we use sentences from the MIMIC-III \cite{25} database and classify
sentences into 4 different classes classes. As for few-shot clinical named entity recognition, we
use the i2b2 (Informatics for Integrating Biology and the Bedside) 2006 de-identitification
challenge dataset \cite{19} and do binary classification of one-word entities. The datasets were
preprocessed in order to be used in few-shot experiments.

\subsubsection*{\textit{MIMIC-III}}

For sentence classification, the sentences were obtained from the MIMIC-III database. We used the
same dataset as in the HealthPrompt paper by Sivarajkumar and Wang \cite{26}, but with classes
suitable for 4-shot, 8-shot, and 16-shot experiments. We got the following classes:
\texttt{ADVANCED.LUNG.DISEASE} (245 samples), \texttt{ADVANCED.HEART.DISEASE} (117 samples),
\texttt{CHRONIC.PAIN.FIBROMYALGIA} (48 samples), and \texttt{ADVANCED.CANCER} (34 samples).

In total, we had 444 samples in our dataset. Since we performed 4, 8, and 16 shot experiments, the
train size varied and was 16, 32, and 64 samples with the test sizes of 428, 412, and 380 samples
respectively. The dataset had two columns: \texttt{text} and \texttt{label} corresponding to the
sentence and label respectively.

\subsubsection*{\textit{2006 i2b2 De-Identification Challenge Dataset}}

The i2b2 (Informatics for Integrating Biology and the Bedside) organized clinical NLP challenge in
2006 with de-identification track focused on identifying Protected Health Information (PHI) from
clinical narratives. The i2b2 2006 challenge developed a corpus of de-identified records such that
there is one record per patient.

The dataset has both training and testing sets, which are separate from each other. This allows for
a more transparent comparison reporting of benchmarks and statistics. One limitation that we also
discuss in the ``Future Work'' section is that we filtered out all multi-word entities and only
perform a one-word NER. We also note this in ``Discussion'' section.

Both training and testing sets are XML files comprised of text, some of which contain \texttt{PHI}
tags. One example of such text is the following:

\begin{center}
\texttt{<PHI TYPE="DATE">11/28</PHI>/03 02:25 PM}
\end{center}

Here, we have text where \texttt{11/28} is a PHI, whose type is \texttt{DATE}. The goal is to
correctly identify this named entity. While the original dataset does contain classes for
\texttt{PHI} tags (e.g., \texttt{DATE} in this case), we do not use them as separate classes for
NER, but only to identify word as a named entity. But similar to sentence classification, we do
sample equal number of samples based on class.

We used two files provided by the challenge:
\texttt{deid\_surrogate\_test\_all\_groundtruth\_version2.xml} and
\texttt{deid\_surrogate\_train\_all\_version2.xml}. All of the \texttt{PHI}s except for \texttt{AGE}
were used. \texttt{AGE} was excluded since the train file contained only 13 such tags, which is not
enough for conducting 16-shot experiments. Finally, we had 8 classes: \texttt{ID},
\texttt{HOSPITAL}, \texttt{DATE}, \texttt{PATIENT}, \texttt{LOCATION}, \texttt{DOCTOR},
\texttt{PHONE}, \texttt{NONPHI} (class that represents non-PHI/non-named-entity words).

We split sentences into words and generated word-level contextual embeddings for every word. For the
training set, we obtained 3\_191 (named entities) and 16\_036 (other terms, labeled
\texttt{NONPHI}). For the testing set, we got 9\_083 (named entities) and 39\_319 (other terms,
labeled \texttt{NONPHI}). While the entire test set was used in all experiments, only 32, 64, and
128 samples were taken from the training set for 4-shot, 8-shot, and 16-shot experiments
respectively. All experiments were run on four NVIDIA RTX 8000 GPUs.


\section*{Methods}

\subsection*{\textit{Sentence-Level Embeddings}}

For contextual, sentence-level embeddings, we used sentence-transformers \cite{24} package. This
package provides a set of intuitive and easy-to-use methods for computing dense vector
representations of sentences, paragraphs, and images. The models are based on transformer networks
such as BERT, RoBERTa \cite{29}, etc., and achieve state-of-the-art performance in various tasks.
The generated sentence embeddings are such that similar texts are close in the latent space and can
efficiently be found using cosine similarity.

\begin{equation}
    \text{cosine similarity(A, B)} = \frac{A \cdot B}{\left\|\mathbf {A} \right\|\left\|\mathbf {B} \right\|}
\end{equation}

\subsection*{\textit{Word-Level Embeddings}}

To generate contextual embeddings at the word level, we used Transformers package \cite{21}.

A tricky part of named entity recognition tasks with transformer models is that they rely on word
piece tokenization, rather than word tokenization. For example, for a word such as
\texttt{Washington}, it may get tokenized into three separate words \texttt{Wash}, \texttt{ing}, and
\texttt{ton}. Then one approach could be to handle this by only training the model on the tag labels
for the first word piece token of a word (i.e. only label ''Wash``). This is what was done in the
original BERT paper.

That being said, such approach does not work well if words that may have prefixes that can also be
tokens. One such example can be a word \texttt{proto-potato}. In this case, tokenizer may split it
into \texttt{proto} and \texttt{potato} and taking the first subword as the embedding of the entire
word would have a semantically incorrect representation. Similarly, if we only took the last subword
embedding, some words may have a suffix and we run into the same problem.

In order to solve this issue, we averaged out the embeddings for the subwords and took it as the
representation for the word. Thus, we define the word embedding \(E_{\text{word}}\) with \(e_0, e_1,
\dots, e_n\) subwords as follows:

\begin{equation}
    E_{\text{word}} = \frac{\sum_{i = 0}^n e_i}{n}
\end{equation}

\subsection*{\textit{Model Architecture}}

\begin{figure}[H]
    \centering
    \includegraphics[width=\linewidth]{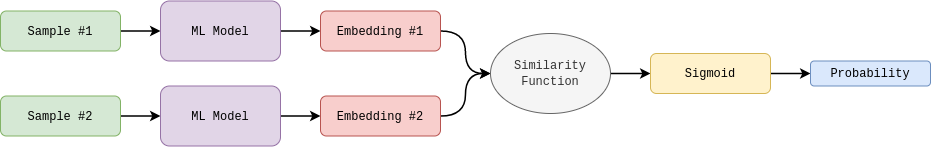}
    \caption{Siamese Neural Network (SNN) Architecture.}
    \label{snn_architecture}
\end{figure}

The SNN's architecture leverages two parallel weight-sharing ML models
(Fig.~\ref{snn_architecture}). In the forward pass, two samples are passed into the models and
mapped down to the latent space. The embeddings in the latent space are then compared using a
similarity function, as shown in (Eq. \ref{equ.sim}). The similarity function is also a parameter
that can be fine-tuned and could range from Euclidean distance to Manhattan distance or cosine
similarity. Depending on the similarity function, the similarity value can then be mapped onto the
\((0, 1)\) interval by applying the Sigmoid function. Finally, a high similarity value means that
the input samples likely belong to the same category, and vice versa.

\begin{equation}
    \label{equ.sim}
    \text{out} = \sigma(\text{distance}(emb_1, emb_2))
\end{equation}

During the training process, SNN conducts representation learning \cite{30} and attempts to have the
best approximation for the input embeddings. The representation is learned by penalizing the loss if
the model yields a high similarity value for inputs from different classes or if the model yields a
low similarity value for inputs from the same class.

The SNN architecture naturally allows for data augmentation. For instance, in the case of 8-shot
learning, the traditional training approach would involve passing 8 samples directly into the model.
This approach is very limiting with a such small number of samples. SNN takes a different route and
instead, it considers unique comparisons within the training set. With the training set comprised of
8 unique samples, there are \(8 * 7\ /\ 2 = 28\) unique comparisons in total. Thus, instead of 8
training samples, we get 28, which is 3.5 times more. In case of 16 samples, the improvement is even
more significant as the number of unique comparisons is 120 and there is 7.5-fold data augmentation.

More generally, under \(N\)-way-\(K\)-shot-classification settings, for the dataset
\(D_\text{train}\) with \(N\) class labels and \(K\) labeled samples for each class, the following
holds after SNN-style augmentation:

\begin{align}
    D_\text{train SNN} &= \{ (x_i, x_j) \mid x_i, x_j \in D_\text{train}, i < j\}\\
    \text{size}(D_\text{train SNN}) &= \frac{(NK)^2 - NK}{2}
\end{align}

Hence, SNN reformulates the classification problem into a pairwise comparison problem, which has the
benefit of obtaining more training data. This becomes increasingly important as the number of
training samples decreases.

\subsection*{\textit{Pre-Trained SNN (PT-SNN)}}

In the first approach, we leverage the pre-trained language models (PLMs) to generate embeddings for
the SNN, called Pre-Trained SNN (PT-SNN). We used three PLMs in this approach, namely BERT, BioBERT,
and BioClinicalBERT, to generate embeddings for the input training samples. In the following, we
illustrate how to use the PT-SNN in the testing step. Suppose we want to perform binary
classification, we are given two classes \(C_1\) and \(C_2\), a training set \(D_\text{train}\), and
a testing set \(D_\text{test}\). We first compute embeddings for all samples in both
\(D_\text{train}\) and \(D_\text{test}\). For every testing sample, using the generated embeddings,
we compute the similarity with respect to every training sample and compute the mean similarity
values for classes.  For instance, mean similarity value for some sample \(x \in D_\text{test}\)
with respect to \(C_1\) and \(C_2\) might be 0.2 and 0.6 respectively. In this case, since 0.6 is
greater than 0.2, we classify sample \(x\) as being in the class \(C_2\). We have also tried using
maximum similarity values per class instead of mean similarity scores, but since the observed
results were similar, we only include results using the mean similarity approach.

\begin{algorithm}[H]
    \caption{Our Proposed Algorithm for SNN-Style Classification and Evaluation for Few-Shot Learning}
    \label{algorithm}
    \begin{algorithmic}[1]
        \Require $D_{train}$: Train dataset
        \Require $E_{test}$: Test dataset embeddings
        \Require $L_{test}$: Test dataset labels
        \Require $d$: Distance function
        \Require $epochs$: Number of evaluation epochs
        \Require $RandSubset$: A function that randomly subsets a dataset with the given seed
            and generates embeddings
        \Require $Mean$: Calculates the mean of a vector
        \Require $GetMaxValueKey$: A function that gets the key with the maximum value from the hash table
        \Require $ComputeMetrics$: A function for computing evaluation metrics -- precision, recall, and F-score
        \State $metrics \gets \text{empty hash table}$;
        \For{$i \gets 0$ to $epochs$}
            \State $E_{train}, L_{train} \gets RandSubset(D_{train}, seed=i)$;
            \State $predictions \gets \text{empty vector}$;
            \ForEach{$e_{test} \in E_{test}$}
                \State $similarity \gets \text{empty hash table}$;
                \ForEach{$(e_{train}, l_{train}) \in Zip(E_{train}, L_{train})$}
                    \State $similarity.Key(l_{train}).Insert(d(e_{train}, e_{test}))$;
                \EndFor
                \State $tmp \gets \text{empty hash table}$;
                \ForEach{$(key,\ val) \in similarity$}
                    \State $tmp.Key(key) = Mean(val)$;
                \EndFor
                \State $predictions.Insert(GetMaxValueKey(tmp))$;
            \EndFor
            \State $precision, recall, fscore = ComputeMetrics(predictions, L_{test})$;
            \State $metrics.Key("precision").Insert(precision)$;
            \State $metrics.Key("recall").Insert(precision)$;
            \State $metrics.Key("fscore").Insert(precision)$;
        \EndFor
        \State $precision = Mean(metrics["precision"]))$;
        \State $recall = Mean(metrics["recall"]))$;
        \State $fscore = Mean(metrics["fscore"]))$;
    \end{algorithmic}
\end{algorithm}

Algorithm \ref{algorithm} presents the pseudocode that contains both the classification algorithm as
well as evaluation approach. Here, \texttt{epochs} refers to the number of averaging epochs for
addressing the instability issues. In our case, \texttt{epochs} is 3. \texttt{d} represents a
distance function, which is cosine similarity in our case. It should be noted that the algorithm is
similar to K-Nearest Neighbors (KNN) \cite{28} classification algorithm.

Such a strategy for classification can be slow in cases where the training set is very large.
However, the proposed approach is feasible in the FSL setting, where the number of annotated samples
is limited. Thus, we do not expect significant performance drawbacks when the number of samples is
not very large.

\subsection*{\textit{SNN With Second-Order Embeddings (SOE-SNN)}}

\begin{figure}[H]
    \centering
    \includegraphics[width=\linewidth]{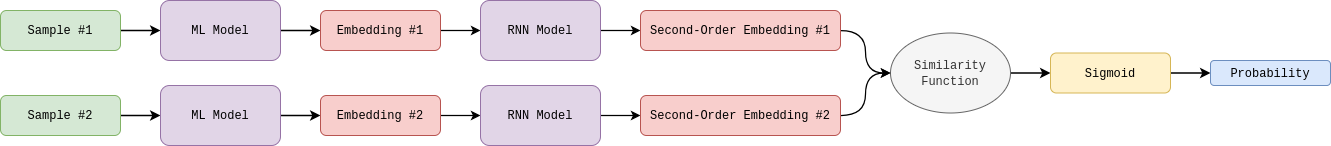}
    \caption{Siamese Neural Network with Second-Order Embeddings (SOE-SNN) Architecture.}
    \label{soe_snn_architecture}
\end{figure}

The second proposed approach is SNNs with second-order embeddings where we apply an additional
Recurrent Neural Network (RNN) layer, such as Long-Short Term Memory (LSTM) or Gated Recurrent Unit
(GRU) to the generated embeddings and then train the SNN model in the fashion described in the Model
Architecture section (Fig.~\ref{soe_snn_architecture}). In our experiments, we used bidirectional
RNNs for producing second-order embeddings.

Specifically, we first obtain the embeddings for all training samples from the PLMs. All possible
unique pairs of samples are then generated and given a label of 0 if the samples in the pair are of
the same class or 1 if they come from different classes. During the training process, we use the
Binary Cross Entropy (BCE) \cite{22} and AdamW \cite{23} as the loss function and the optimizer,
respectively.

Model evaluation is done in the same manner as in pre-trained SNNs, where we compute mean similarity
scores and average out the metrics over 3 testing epochs to handle the potential instability issues.

\subsection*{\textit{FSL Model Evaluation}}

Systematically evaluating FSL model performance can be tricky since fine-tuning or making
predictions on small datasets could potentially suffer from instability \cite{20}. To address this
issue, we propose the averaging strategy for model evaluation. For every few-shot experiment (e.g.,
4-shot, 8-shot, and 16-shot experiments), using randomized sampling, we sample 4, 8, or 16 samples
per class and create a training dataset. We perform this \(M\) times and therefore, for every
experiment, \(M\) randomly generated training sets are evaluated on the test set. Finally, the
metrics are averaged out and reported as the final scores.

\begin{equation}
    \text{Metric} = \frac{\sum_{i = 1}^M \text{Metric}_i}{M}
\end{equation}

In this paper, we choose \(M = 3\). Such approach gives a more robust view on the performance of the
model in possibly unstable scenarios. We employ this strategy in all reported metrics. As for
metrics, we choose precision, recall, and F-score.

\subsection*{\textit{Baseline Models}}

We use fine-tuned BERT, BioBERT, and BioClinicalBERT as the baseline models. Instead of approaching
the problem from the SNN perspective, we use the 4, 8, and 16 samples per class directly in order to
fine-tune the pre-trained transformer models.

This is done by adding an additional linear layer at the end of the transformer model. To achieve
this, we, once again, used Transformers package \cite{21}.


\section*{Results}

We present the results of 4-shot, 8-shot, and 16-shot experiments for few-shot text classification
tasks -- Sentence Classification (SC) and Named-Entity Recognition (NER). We used models based on
BERT, BioBERT, BioClinicalBERT. The results are shown in Table 1. Note that PT-SNN, SOE-SNN, and FTT
stand for Pre-Trained Siamese Neural Network, Siamese Neural Network with Second-Order Embeddings,
and Fine-Tuned Transformer respectively.

\begin{table}[H]
    \begin{center}
        \begin{tabular}{|l|l|c|c|c|c|}
            \hline
            \textbf{Approach} & \textbf{Model} & \textbf{Shots} & \textbf{Precision (SC / NER)} & \textbf{Recall (SC / NER)} & \textbf{F-Score (SC / NER)}\\
            \hline

            FTT & BERT & 4 & 0.24 / 0.70 & 0.24 / 0.59 & 0.14 / 0.60\\
            FTT & BioBERT & 4 & 0.23 / 0.66 & 0.23 / 0.56 & 0.16 / 0.57\\
            FTT & BioClinicalBERT & 4 & 0.25 / 0.66 & 0.31 / 0.19 & 0.18 / 0.15\\

            PT-SNN & BERT & 4 & 0.49 / 0.87 & 0.37 / 0.45 & 0.37 / 0.49\\
            PT-SNN & BioBERT & 4 & \textbf{0.53} / 0.87 & \textbf{0.45} / 0.24 & \textbf{0.46} / 0.18\\
            PT-SNN & BioClinicalBERT & 4 & 0.50 / \textbf{0.89} & 0.42 / \textbf{0.64} & 0.43 / \textbf{0.68}\\

            SOE-SNN & BERT & 4 & 0.47 / 0.80 & 0.51 / 0.42 & 0.42 / 0.41\\
            SOE-SNN & BioBERT & 4 & 0.44 / 0.83 & 0.47 / 0.48 & 0.44 / 0.50\\
            SOE-SNN & BioClinicalBERT & 4 & 0.50 / 0.76 & 0.47 / 0.58 & 0.42 / 0.61\\

            \hline

            FTT & BERT & 8 & 0.34 / 0.74 & 0.30 / 0.52 & 0.24 / 0.53\\
            FTT & BioBERT & 8 & 0.35 / 0.70 & 0.32 / 0.39 & 0.22 / 0.38\\
            FTT & BioClinicalBERT & 8 & 0.35 / 0.71 & 0.35 / 0.30 & 0.21 / 0.29\\

            PT-SNN & BERT & 8 & 0.62 / 0.87 & 0.45 / 0.37 & 0.47 / 0.37\\
            PT-SNN & BioBERT & 8 & 0.61 / 0.87 & 0.48 / 0.25 & 0.50 / 0.19\\
            PT-SNN & BioClinicalBERT & 8 & 0.64 / \textbf{0.88} & 0.44 / \textbf{0.55} & 0.49 / \textbf{0.59}\\

            SOE-SNN & BERT & 8 & 0.59 / 0.88 & 0.47 / 0.35 & 0.50 / 0.36\\
            SOE-SNN & BioBERT & 8 & 0.62 / 0.87 & 0.52 / 0.24 & 0.55 / 0.20\\
            SOE-SNN & BioClinicalBERT & 8 & \textbf{0.65} / 0.88 & \textbf{0.51} / 0.54 & \textbf{0.55} / 0.60\\
            \hline
            FTT & BERT & 16 & 0.23 / 0.76 & 0.31 / 0.54 & 0.14 / 0.55\\
            FTT & BioBERT & 16 & 0.35 / 0.71 & 0.33 / 0.40 & 0.18 / 0.39\\
            FTT & BioClinicalBERT & 16 & 0.39 / 0.74 & 0.37 / 0.37 & 0.27 / 0.34\\

            PT-SNN & BERT & 16 & 0.64 / 0.87 & 0.51 / 0.33 & 0.52 / 0.32\\
            PT-SNN & BioBERT & 16 & 0.65 / 0.87 & 0.55 / 0.25 & 0.56 / 0.20\\
            PT-SNN & BioClinicalBERT & 16 & 0.69 / 0.88 & 0.54 / 0.47 & 0.58 / 0.49\\

            SOE-SNN & BERT & 16 & 0.66 / 0.88 & 0.58 / 0.34 & 0.58 / 0.30\\
            SOE-SNN & BioBERT & 16 & 0.68 / 0.88 & 0.55 / 0.27 & 0.56 / 0.22\\
            SOE-SNN & BioClinicalBERT & 16 & \textbf{0.71} / \textbf{0.90} & \textbf{0.55} / \textbf{0.50} & \textbf{0.60} / \textbf{0.51}\\
            \hline
        \end{tabular}
        \caption{Few-Shot Sentence Classification (SC) and Named-Entity Recognition (NER) Results.}
    \end{center}
\end{table}

\subsection*{\textit{Few-Shot Clinical Text Classification}}

In the 4-shot sentence classification task, the baseline, fine-tuned transformer, approach had the
worst performance. PT-SNN and SOE-SNN both significantly outperformed fine-tuned transformer, with
PT-SNN marginally outperforming SOE-SNN. The best model was BioBERT-based PT-SNN, with the precision
of 0.53, recall of 0.45, and F-score of 0.46. The best FTT model was based on BioClinicalBERT and
had the precision of 0.25, recall of 0.31, and F-score of 0.18. In order to measure the statistical
significance of the difference between the best fine-tuned transformer approach
(BioClinicalBERT-based model) and the best SNN-based approach, we perform the paired t-test. This is
done by creating two vectors \([\text{precision}_1, \text{recall}_1, \text{fscore}_1]\) and
\([\text{precision}_2, \text{recall}_2, \text{fscore}_2]\), calculating the difference vector \(d\),
and then running one-sample t-test. Difference was significant with the p-value of 0.0377 (we take
the 0.05 cutoff).

In 8-shot experiments, BioClinicalBERT-based SOE-SNN outperformed all other approaches, with the
precision, recall, and F-score of 0.65, 0.51, and 0.55 respectively. PT-SNN came next with
fine-tuned transformers having the worst performance. The best baseline model (BioClinicalBERT-based
model) had the precision, recall, and F-score values of 0.35, 0.35, and 0.21 respectively. After
performing the paired t-test, we got the p-value of 0.0394. Thus, the difference is significant and
the proposed approach is a substantial improvement over the baseline models.

As for 16-shot learning, BioClinicalBERT-based SOE-SNN outperformed all other models, with the
precision, recall, and F-score of 0.71, 0.55, and 0.60 respectively. The worst performance, once
again, was shown by the baseline (BioClinicalBERT-based) fine-tuned transformer model, with
precision, recall, and F-score of 0.39, 0.37, and 0.27 respectively. After performing the paired
t-test, we obtained the p-value of 0.0293 and hence, we conclude that the difference between the
metrics is significant and therefore, the proposed approach significantly outperforms the baseline
models.

Overall, it is clear that SNN-based approaches outperformed the baseline models and the difference
in performance was significant in all cases. PT-SNN and SOE-SNN performed similarly and the
difference in their performance was not statistically significant. It should also be noted that we
measured statistical significance on the metrics directly and not on model outputs.

\subsection*{\textit{Few-Shot Clinical Named Entity Recognition}}

In 4-shot experiments, SNN-based model (BioClinicalBERT-based PT-SNN) had the best performance with
the precision, recall, and F-score of 0.89, 0.64, and 0.68 respectively. The best baseline model was
the BERT-based fine-tuned transformer with the precision of 0.70, recall of 0.59, and F-score of
0.60. The paired t-test got yielded the p-value of 0.1291, which is not statistically significant.

In 8-shot settings, similar to the 4-shot scenario, BioClinicalBERT-based PT-SNN was the best
model. The best baseline model was the BERT-based fine-tuned transformer. SNN-based model achieved
the precision, recall, and F-score of 0.88, 0.55, and 0.59 respectively, while the fine-tuned
transformer had those of 0.74, 0.52, and 0.53 respectively. The paired t-test got the p-value of
0.1446, which is not statistically significant.

As for 16-shot NER task, BioClinicalBERT-based SOE-SNN outperformed all other models, with
precision, recall, and F-score of 0.90, 0.50, and 0.51 respectively. The best baseline model was the
BERT-based fine-tuned transformer, which achieved the precision, recall, and F-score of 0.76, 0.54,
and 0.55 respectively. The paired t-test got us the p-value of 0.7706, which is not statistically
significant.

Overall, despite the results not being statistically significant, SNN-based approaches showed a
noticeable performance improvement over fine-tuned transformers. It should also be noted that as the
number of samples increased, metrics recall and F-score have decreased for the SNN-based approaches.
This pattern was not observed in few-shot sentence classification tasks and could potentially be
explained by lack of sufficient context in named entities and the number of named entities.


\section*{Discussion}

There are several limitations of the work that can be addressed by further exploring few-shot
learning and SNNs. First, one could compare the results to more traditional baseline models such as
SVM, logistic regression, multinomial logistic regression, random forest, etc. Second, other
datasets could also be used for evaluating the performance of SNNs in text classification. Third, we
had a limitation when doing NER -- particularly, we only considered one-word entities. Evaluating
the performance of SNNs considering all entities could be interesting. Additionally, since we can
perform both word-level and sentence-level classification, another interesting direction of research
could be document classification, where a document is a collection of words and sentences.
Furthermore, it is important to note that datasets for FSL and especially, for clinical FSL are
difficult to find. Ge et al. \cite{27}, in their paper, have emphasized that ``(68\%) studies
reconstructed existing datasets to create few-shot scenarios synthetically.'' Hence, building the
brand new FSL dataset and then evaluating the performance of proposed methods could also be an
interesting future research direction.


\section*{Conclusion}

We have conducted few-shot learning experiments evaluating the performance of SNN models on text
classification tasks -- SC and NER. SNN models were based on transformer models -- BERT, BioBERT,
and BioClinicalBERT. Fine-tuned variants of  BERT, BioBERT, and BioClinicalBERT were used as the
baseline models. Since performance evaluation on small datasets may suffer from instability, a
special evaluation strategy was used. We conclude that SNN-based models outperformed the baseline
fine-tuned transformer models for sentence classification tasks. The paired t-test was also
performed, which showed that for SC tasks, SNN models significantly outperformed the baseline
models. As for NER, SNN models, once again, outperformed the baseline models, yet the performance
difference was not statistically significant. The limitations of the work have also been discussed
alongside with potential future directions of research.


\subparagraph{Acknowledgments}

The authors would like to acknowledge support from the University of Pittsburgh Momentum Funds,
Clinical and Translational Science Institute Pilot Awards, the
School of Health and Rehabilitation Sciences Dean’s Research and Development Award, and the National
Institutes of Health through Grant UL1TR001857.

\makeatletter
\renewcommand{\@biblabel}[1]{\hfill #1.}
\makeatother

\bibliographystyle{unsrt}
\bibliography{amia}

\end{document}